\documentclass{article}

\usepackage{microtype}
\usepackage{graphicx}
\usepackage{subfigure}
\usepackage{booktabs}
\usepackage{hyperref}
\usepackage{multirow}
\usepackage{pifont}

\usepackage[accepted]{icml2019}

\icmltitlerunning{An Empirical Evaluation on Robustness and Uncertainty of Regularization Methods}

\begin{document}

\twocolumn[
\icmltitle{An Empirical Evaluation on\\Robustness and Uncertainty of Regularization Methods}

\icmlsetsymbol{equal}{*}

\begin{icmlauthorlist}
\icmlauthor{Sanghyuk Chun}{clair}
\icmlauthor{Seong Joon Oh}{line}
\icmlauthor{Sangdoo Yun}{clair}
\icmlauthor{Dongyoon Han}{clair}
\icmlauthor{Junsuk Choe}{yonsei}
\icmlauthor{Youngjoon Yoo}{clair}
\end{icmlauthorlist}

\icmlaffiliation{clair}{Clova AI Research, NAVER Corp.}
\icmlaffiliation{line}{Clova AI Research, LINE Plus Corp.}
\icmlaffiliation{yonsei}{Yonsei University}

\icmlcorrespondingauthor{Sanghyuk Chun}{sanghyuk.c@navercorp.com}

\icmlkeywords{Robustness, Uncertainty, Regularization, Machine Learning, ICML}

\vskip 0.3in
]

\printAffiliationsAndNotice{}

\begin{abstract}
Despite apparent human-level performances of deep neural networks (DNN), they behave fundamentally differently from humans. They easily change predictions when small corruptions such as blur and noise are applied on the input (lack of robustness), and they often produce confident predictions on out-of-distribution samples (improper uncertainty measure). While a number of researches have aimed to address those issues, proposed solutions are typically expensive and complicated (e.g. Bayesian inference and adversarial training). Meanwhile, many simple and cheap regularization methods have been developed to enhance the generalization of classifiers. Such regularization methods have largely been overlooked as baselines for addressing the robustness and uncertainty issues, as they are not specifically designed for that. In this paper, we provide extensive empirical evaluations on the robustness and uncertainty estimates of image classifiers (CIFAR-100 and ImageNet) trained with state-of-the-art regularization methods. Furthermore, experimental results show that certain regularization methods can serve as strong baseline methods for robustness and uncertainty estimation of DNNs.
\end{abstract}

\section{Introduction}
Recent studies have shown that inner mechanisms of DNNs are different from those of humans. For example, DNNs are easily fooled by human-imperceptible adversarial perturbations (adversarial robustness \cite{szegedy2013intriguing, fgsm}) and semantics-preserving transformations like noising, blurring, and texture corruptions (natural robustness \cite{geirhos2018generalisation, hendrycks2018imagenet-c, geirhos2018imagenet_stylize}). Another limitation of DNNs is their inability to produce sound uncertainty estimates for their predictions. They are known to be inept at producing well-calibrated predictive uncertainties (known unknowns) and detecting out-of-distribution (OOD) samples (unknown unknowns) \cite{hendrycks2016baseline}.

For adversarial robustness, it has been shown that augmenting adversarial perturbations during training, or adversarial training, makes a model more adversarially robust \cite{kurakin2016adversarial, madry2017towards, xie2018feature}. However, it is computationally challenging to employ it on large-scale datasets \cite{kurakin2016adversarial, xie2018feature}. Adversarially trained models overfit to the specific attack type used for training \cite{sharma2017attacking}, and the performance on unperturbed images drops \cite{tsipras2018robustness}. On the other hand, methods which improve robustness to non-adversarial corruptions are relatively less studied. Recently it is shown that training models by augmenting a specific noise enhances the performance on the target noise but can not be generalized to the other unseen noise types \cite{geirhos2018generalisation}. ImageNet-C dataset \cite{hendrycks2018imagenet-c} is proposed to evaluate robustness to $15$ corruption types including blur and noise while a network should not observe the distortions during the train time. The authors have shown that the natural robustness is improved via adversarial training \cite{ALP} and Stylized ImageNet augmentation \cite{geirhos2018imagenet_stylize}, but have not considered more common and simpler regularization types; we provide those baseline experiments in this paper.

Efforts to improve uncertainty estimates of DNNs have followed two distinguishable paths: improving calibration of predictive uncertainty and out-of-distribution (OOD) sample detection. On the predictive uncertainty side, variants of Bayesian neural networks \cite{MCDropout, UncertaintyTypes} and ensemble methods \cite{EnsembleUncertainty} have mainly been proposed. These approaches, however, are expensive and often require modifications of training and inference stages. On the OOD detection front, methods including threshold-based binary classifiers \cite{hendrycks2016baseline} and real or GAN-generated OOD sample augmentation \cite{lee2018confident} have brought about improvements in OOD detections. Above approaches have demonstrated sub-optimal performances in our experiments, even compared to simple baselines.

As an independent line of research, many regularization techniques have been proposed to improve the generalization of DNN classifiers. For example, Batch Normalization (BN) \cite{BN} and data augmentation strategies such as random crop and random flip \cite{alexnet, InceptionResnet} have become standard design choices for deep models. Despite their simplicity and efficiency, the effects of state-of-the-art regularization techniques such as label smoothing \cite{szegedy2016rethinking_labelsm}, MixUp \cite{zhang2017mixup} and CutMix \cite{yun2019cutmix} on the robustness and the uncertainty of deep models are still rarely investigated. A few works have shown indeed the effects of a few regularization techniques on DNN robustness \cite{zhang2017mixup, ALP, yun2019cutmix}, but we provide a more extensive analysis with both robustness and uncertainty perspectives.

We empirically evaluate state-of-the-art regularization techniques and show that they improve the classification, robustness, and uncertainty estimates for large-scale classifiers at marginal additional costs. We argue that certain regularization techniques must be considered as strong baselines for future researches in robustness and uncertainty of DNNs.

\newcommand\best[1]{\bf \textcolor{red}{ #1}}
\newcommand\second[1]{\bf \textcolor{blue}{ #1}}
\newcommand\third[1]{\bf \textcolor{brown}{ #1}}

\newcommand\truemark{\ding{51}}
\newcommand\falsemark{$-$}

\begin{table*}[ht!]
\centering
\small
\caption{CIFAR-100 classification, robustness to adversarial and non-adversarial noises, and uncertainty benchmark results. For non-adversarial corruptions, we report top-1 error in CIFAR-100-C dataset and top-1 error in occluded CIFAR-100 test samples. We report out-of-distribution (OOD) detection errors averaged over seven OOD datasets. We fix the base architecture as PyramidNet-200 with $\alpha=240$. LS stands for label smoothing. Lower is better for all reported numbers and all values are percentage.}
\label{table:overview}
\vspace{0.1cm}
\begingroup
\renewcommand{\arraystretch}{1.12}
\begin{tabular}{lcccccccccc}
                                  &              &  & Classification &  & \multicolumn{3}{c}{Robustness}          &  & \multicolumn{2}{c}{Uncertainty} \\ \cline{4-4} \cline{6-8} \cline{10-11} 
                                  &              & & CIFAR-100      &  & FGSM        & CIFAR-C & Occlusion   &  & Expected  & OOD              \\
Method                            & LS           & & Top-1 Err.    &  & Top-1 Err. & Top-1 Err.         & Top-1 Err. &  & Calibration Err.  & Detection Err.  \\ \cline{1-2} \cline{4-4} \cline{6-8} \cline{10-11} 
\multirow{2}{*}{Baseline}         & \falsemark{} &  & 16.45          &  & 84.20       & 45.11       & 72.19       &  & 8.00         & 18.05            \\
                                  & \truemark{}  &  & 16.73          &  & 82.82       & 46.50       & 74.40       &  & 2.51         & 17.59            \\ \cline{1-2} \cline{4-4} \cline{6-8} \cline{10-11}
\multirow{2}{*}{ShakeDrop}        & \falsemark{} &  & 15.08          &  & 77.91       & 44.37       & 78.69       &  & 8.01         & 19.76            \\
                                  & \truemark{}  &  & 15.05          &  & 63.09       & 43.74       & 82.22       &  & 2.53         & 25.59            \\ \cline{1-2} \cline{4-4} \cline{6-8} \cline{10-11}
\multirow{2}{*}{Cutout}           & \falsemark{} &  & 16.53          &  & 91.07       & 51.65       & 27.00       &  & 7.67         & 28.73            \\
                                  & \truemark{}  &  & 15.61          &  & 77.77       & 48.74       & 27.03       &  & 4.24         & 17.92            \\ \cline{1-2} \cline{4-4} \cline{6-8} \cline{10-11}
\multirow{2}{*}{Cutout + ShakeDrop}           & \falsemark{} &  & 15.91          &  & 88.66       & 50.00       & 26.19       &  & 6.63         & 19.55            \\
                                  & \truemark{}  &  & 13.49          &  & 69.59       & 43.86       & 26.33       &  & 1.45         & 18.40            \\ \cline{1-2} \cline{4-4} \cline{6-8} \cline{10-11}
\multirow{2}{*}{Mixup}            & \falsemark{} &  & 15.63          &  & 63.85       & 42.81       & 56.80       &  & 7.89         & 39.09            \\
& \truemark{}  &  & 15.91          &  & 55.84       & 42.20       & 57.60       &  & 15.20        & 28.56            \\ \cline{1-2} \cline{4-4} \cline{6-8} \cline{10-11}
\multirow{2}{*}{Mixup + ShakeDrop}            & \falsemark{} &  & 14.91          &  & 61.91       & 40.60       & 57.07       &  & 7.28         & 22.92            \\ 
                                  & \truemark{}  &  & 14.79          &  & 56.32       & 40.32       & 56.76       &  & 15.85        & 18.54            \\ \cline{1-2} \cline{4-4} \cline{6-8} \cline{10-11}
\multirow{2}{*}{CutMix}           & \falsemark{} &  & 14.23          &  & 88.88       & 49.83       & 32.16       &  & 4.92         & 10.95            \\
                                  & \truemark{}  &  & 15.55          &  & 74.00       & 51.01       & 35.68       &  & 7.91         & 13.56            \\ \cline{1-2} \cline{4-4} \cline{6-8} \cline{10-11}
\multirow{2}{*}{CutMix + ShakeDrop}           & \falsemark{} &  & 13.81          &  & 70.75       & 43.36       & 35.83       &  & 2.46         & 19.82            \\
                                  & \truemark{}  &  & 13.83          &  & 62.72       & 44.99       & 34.96       &  & 5.26         & 18.89            \\ \cline{1-2} \cline{4-4} \cline{6-8} \cline{10-11} \vspace{-1em} \\ \cline{1-2} \cline{4-4} \cline{6-8} \cline{10-11}
Adversarial Logit Pairing               & \truemark{}  &  & 24.75          &  & 51.32       & 50.04       & 92.27       &  & 6.67         & 21.57            \\
Adversarial Training                     & \truemark{}  &  & 26.85          &  & 51.80       & 51.85       & 93.59       &  & 8.71         & 28.06            \\ \cline{1-2} \cline{4-4} \cline{6-8} \cline{10-11}
w/o Random Crop \& Flip      & \falsemark{} &  & 21.83          &  & 90.63       & 48.71       & 77.46       &  & 7.99         & 26.91            \\
Add Gaussian Noise                & \falsemark{} &  & 19.49          &  & 85.08       & 42.01       & 73.23       &  & 9.79         & 25.16            \\
OOD augment (SVHN)                        & \falsemark{} &  & 38.80          &  & 97.35       & 67.03       & 79.13       &  & 46.37        & 43.53            \\
OOD augment (GAN)                         & \falsemark{} &  & 34.78          &  & 94.65       & 57.09       & 85.30       &  & 38.22        & 33.35  \\ \cline{1-2} \cline{4-4} \cline{6-8} \cline{10-11}
\end{tabular}
\endgroup
\end{table*}

\vspace{-0.5em}
\section{Revisiting Regularization Methods}

In this section, we revisit several regularization methods including the state-of-the-art regularization methods used in our experiments.

\vspace{-0.5em}
\paragraph{Input augmentation:}
With proper data augmentation methods, a model can generalize better to the unseen samples. For example, random cropping and flipping are widely used to improve classification performances \cite{alexnet, InceptionResnet, densenet}. However, it is not always straightforward to distinguish augmentation types that improves the generalizability. For example, adversarial samples, geometric transformations, and pixel inversion are rarely helpful for improving classification performances \cite{tsipras2018robustness, cubuk2018autoaugment}. One of the most effective augmentation methods is Mixup \cite{zhang2017mixup} which generates the in-between class samples by the pixel level interpolation. Another example of data augmentation is Cutout which erases pixels in a region sampled at random \cite{devries2017cutout, zhong2017randomerase}. Recently proposed CutMix fills the pixels from other images instead of erasing pixels \cite{yun2019cutmix}. While being simple and efficient, Mixup, Cutout and CutMix have shown significant improvements in classification performance. We consider their contribution to robustness and uncertainty estimates in our experiments.

\paragraph{Label perturbation:}
Deep models often suffer from over-confident predictions; they often produce predictions with high confidence even on random Gaussian noise input \cite{hendrycks2016baseline}. One straightforward way to mitigate the issue is to penalize over-confident predictions by perturbing the target $y$. For example, label smoothing \cite{szegedy2016rethinking_labelsm} changes ground-truth label to a smoothed distribution whose probability of non-targeted labels are $\alpha/K$, where $\alpha$ is a smoothing parameter whose default value is often $0.1$ and $K$ is the number of classes. By smoothing target predictions, models learn to regularize overconfident predictions. Another examples are Mixup \cite{zhang2017mixup} and CutMix \cite{yun2019cutmix} which blend two one-hot labels into one smooth label by the mix ratio. Label smoothing is also known to offer a modest amount of robustness to adversarial perturbations \cite{ALP}. It is thus widely used in adversarial training to achieve better adversarial robustness. We consider label smoothing as one of the axes for our investigation.

\paragraph{Other strategies for deep networks:}
Many researches have achieved more stable convergence and better generalization performance via weight regularization (weight decay) or feature-level manipulations like dropout \cite{Dropout} and Batch Normalization \cite{BN}. Recently, randomly adding noises on intermediate features \cite{ghiasi2018dropblock, shakeshake, stochasticdepth, yamada2018shakedrop}, or adding extra paths to the model \cite{SENet, GENet} have been proposed. We present robustness and uncertainty experiments on a selection of above regularization techniques.

\section{Benchmarks for Robustness and Uncertainty Estimation}

In this section, we describe the settings for the benchmarks used in our experiments. We tested four benchmarks: robustness to adversarial attacks, robustness to natural corruptions, robustness to occlusions, confidence calibration error, and out-of-distribution detection.

To evaluate adversarial robustness, we use FGSM \cite{fgsm} with $\epsilon = 8/255$. Note that our baseline regularization methods cannot provide a provable defense to the adversarial attacks while adversarial training and ALP could mitigate the effect of the adversarial attacks.

For evaluating robustness against natural corruptions, we employ naturally corrupted ImageNet (ImageNet-C) proposed by \cite{hendrycks2018imagenet-c}. In ImageNet-C, there are $15$ transforms categorized into ``noise'', ``blur'', ``weather'', and ``digital'' with five severities. For CIFAR-100 experiments, we create corrupted CIFAR-100 (CIFAR-100-C) using $75$ transforms proposed in ImageNet-C. We report the average accuracy over all $75$ transforms.

In occlusion robustness benchmarks, we generate occluded samples by filling zeros (black pixels) over a square at the image center whose side length is half the image width; i.e., $16$ for CIFAR-100 and $112$ for ImageNet.

To show how the methods affect the confidence of predictions, we evaluate the expected calibration error \cite{guo2017calibration}. We view a classification system as a probabilistic confidence estimator whose confidence is a measurement of the trustworthy estimation. The bin size is set to $20$. We refer \cite{guo2017calibration} for further details of the evaluation.

Finally, we have tested the baseline OOD detection performance of each model. We have used the threshold-based detector proposed in \cite{hendrycks2016baseline}. Seven datasets used in \cite{liang2017odin} were considered: cropped Tiny ImageNet, resized Tiny ImageNet, cropped LSUN \cite{yu2015lsun}, resized LSUN, iSUN, Gaussian noise, and Uniform noise. We report the average detection error over the seven datasets.

\vspace{-1em}
\section{Main Results}

\begin{table*}[ht!]
\centering
\small
\caption{Comparison of noise augmentations on robustness to various noises. Noise, blur, weather and digital are a subset of CIFAR-C.}
\label{table:robustness_methods}
\begin{tabular}{@{}lcccc|cccc@{}}
\toprule
             & CIFAR-100  & FGSM       & Occlusion  & CIFAR-C & Noise & Blur & Weather & Digital \\
Methods      & Top-1 Err. & Top-1 Err. & Top-1 Err. & mCE     & Top-1 Err.      & Top-1 Err.     & Top-1 Err.        & Top-1 Err.        \\ \midrule
Baseline     & \best{16.45}      & 84.20      & 72.19      & 45.11   & 74.62           & \best{46.77}          & \best{30.66}             & 38.65             \\
Adversarial Logit Pairing          & 24.75      & \best{51.32}      & 92.27      & 50.04   & 69.94           & 51.75          & 40.62             & 44.70             \\
Cutout       & 16.53      & 91.07      & \best{27.00}      & 51.65   & 89.77           & 51.40          & 34.24             & 43.20             \\
Add Gaussian Noise & 19.49      & 85.08      & 73.23      & \best{42.01}   & \best{54.63}           & 48.42          & 31.54             & \best{38.48}             \\ \bottomrule
\end{tabular}
\end{table*}

\begin{table*}[ht!]
\centering
\small
\vspace{-1em}
\caption{Comparison of well-regularized networks and baseline methods to improve robustness and uncertainty. SD stands for ShakeDrop.}
\label{table:powerful_baseline}
\begin{tabular}{lccccccccc}
\toprule
                                  &              & CIFAR-100      &  & FGSM        & CIFAR-C & Occlusion   &  & Expected  & OOD              \\
Method                            & & Top-1 Err.    &  & Top-1 Err. & Top-1 Err     & Top-1 Err. &  & Calibration Err. & Detection Err.  \\ \midrule
Baseline         & & 16.45          &  & 84.20       & 45.11       & 72.19       &  & 8.00         & 18.05            \\
Cutout + SD + LS                  &  & 13.49          &  & 69.59       & 43.86       & 26.33       &  & 1.45         & 18.40            \\
Mixup + SD + LS           & & 14.79          &  & 56.32       & 40.32       & 56.76       &  & 15.85         & 18.54            \\
CutMix + SD + LS           & & 13.83          &  & 62.72       & 44.99       & 34.96       &  & 5.26         & 18.89            \\ \midrule
Adversarial Logit Pairing               & & 24.75          &  & 51.32       & 50.04       & 92.27       &  & 6.67         & 21.57            \\ 
Add Gaussian Noise                &  & 19.49          &  & 85.08       & 42.01       & 73.23       &  & 9.79         & 25.16            \\
OOD augment (SVHN)                        &  & 38.80          &  & 97.35       & 67.03       & 79.13       &  & 46.37        & 43.53            \\
OOD augment (GAN)                         &  & 34.78          &  & 94.65       & 57.09       & 85.30       &  & 38.22        & 33.35  \\
\bottomrule
\end{tabular}
\vspace{-1em}
\end{table*}

\subsection{Training Settings}

We first describe the settings for training models used in the robustness and the uncertainty benchmarks. To ensure the effectiveness of each regularization methods, we employ a powerful baseline, PyramidNet-200~\cite{pyramidnet} and ResNet-50~\cite{resnet} for CIFAR-100 and ImageNet experiments, respectively.

We consider the state-of-the-art regularization methods of Cutout \cite{devries2017cutout}, Mixup \cite{zhang2017mixup}, CutMix \cite{yun2019cutmix}, label smoothing \cite{szegedy2016rethinking_labelsm}, ShakeDrop \cite{yamada2018shakedrop}, and their combinations for experiments. We optimize the models with the SGD with momentum. We set the batch size to $64$ and training epochs to $300$. The learning rate is initially set to $0.25$ and is decayed by the factor of $1/10$ at $150^{\mathrm{th}}$ and $225^{\mathrm{th}}$ epochs. We also employ random crop and random flip augmentations for all methods, unless specified otherwise.

For the comparison methods for adversarial robustness, we train the baseline model with adversarial training \cite{kurakin2016adversarial, madry2017towards} and adversarial logit paring (ALP) \cite{ALP}. We use Fast Gradient Sign Method (FGSM) \cite{fgsm} with $\epsilon=8/255$ as the threat model. All the results are evaluated with applying label smoothing to achieve better performances. We mix the clean and adversarial samples with the same ratio as proposed in \cite{kurakin2016adversarial}. The optimizer for adversarial training is ADAM \cite{kingma2014adam}.

As the baseline method for CIFAR-C, we consider Gaussian noise augmentation; the same type of perturbation taken from the CIFAR-C dataset \cite{hendrycks2018imagenet-c}. For out-of-distribution (OOD) detection baseline, we augment OOD samples and the target labels to be the uniform label as proposed in \cite{lee2018confident}. We augment two types of OOD samples used in \cite{lee2018confident}: Street View House Numbers (SVHN) dataset and generated samples by GAN. In our experiments, we use WGAN-GP \cite{wgangp} instead of DC-GAN \cite{dcgan}.

All the experiments are done with NAVER Smart Machine Learening (NSML)~\cite{nsml1, nsml2}.

\subsection{CIFAR-100 Results}

In this section, we evaluate the effects of the state-of-the-art regularization techniques on the various robustness and uncertainty benchmarks on CIFAR-100. We show that well-regularized models are powerful baselines.

In Table~\ref{table:overview}, we report classification, adversarial and natural robustness, and uncertainty measure evaluations. Classification performances are measured on CIFAR-100 test set; adversarial robustness is measured against the FGSM \cite{fgsm} attack on CIFAR-100; natural robustness is measured on CIFAR-C \cite{hendrycks2018imagenet-c}. Uncertainty qualities are measured in terms of expected calibration error \cite{guo2017calibration} and OOD detection error rates \cite{hendrycks2016baseline}. We report the OOD detection errors at method-specific optimal thresholds.

Here we analyze the following questions from Table~\ref{table:overview}.

\textit{Can data augmentation improve robustness against various perturbations at once?} Data augmentation is a straightforward solution to improve robustness against specific type of noise, e.g., adversarial perturbation, Gaussian noise, and occlusion. In Table~\ref{table:robustness_methods}, we have observed that different type of augmentation methods improve robustness against the target noise. For example, ALP improves adversarial robustness but it fails to improve robustness against occlusion and other natural corruptions. Similarly, in Table~\ref{table:robustness_methods}, Cutout is only method that improves occlusion robustness among the other augmentation methods. However, Cutout degrades other types of robustness, such as adversarial robustness, compare to the baseline. By adding Gaussian noise to the input, robustness to the common corruptions is enhanced, especially for the ``noise''. In summary, we have observed that it would be difficult to improve the robustness against various type of corruptions at once. A similar phenomenon was also observed by \cite{geirhos2018generalisation}.

\begin{figure}[h]
    \centering
    \includegraphics[width=\linewidth]{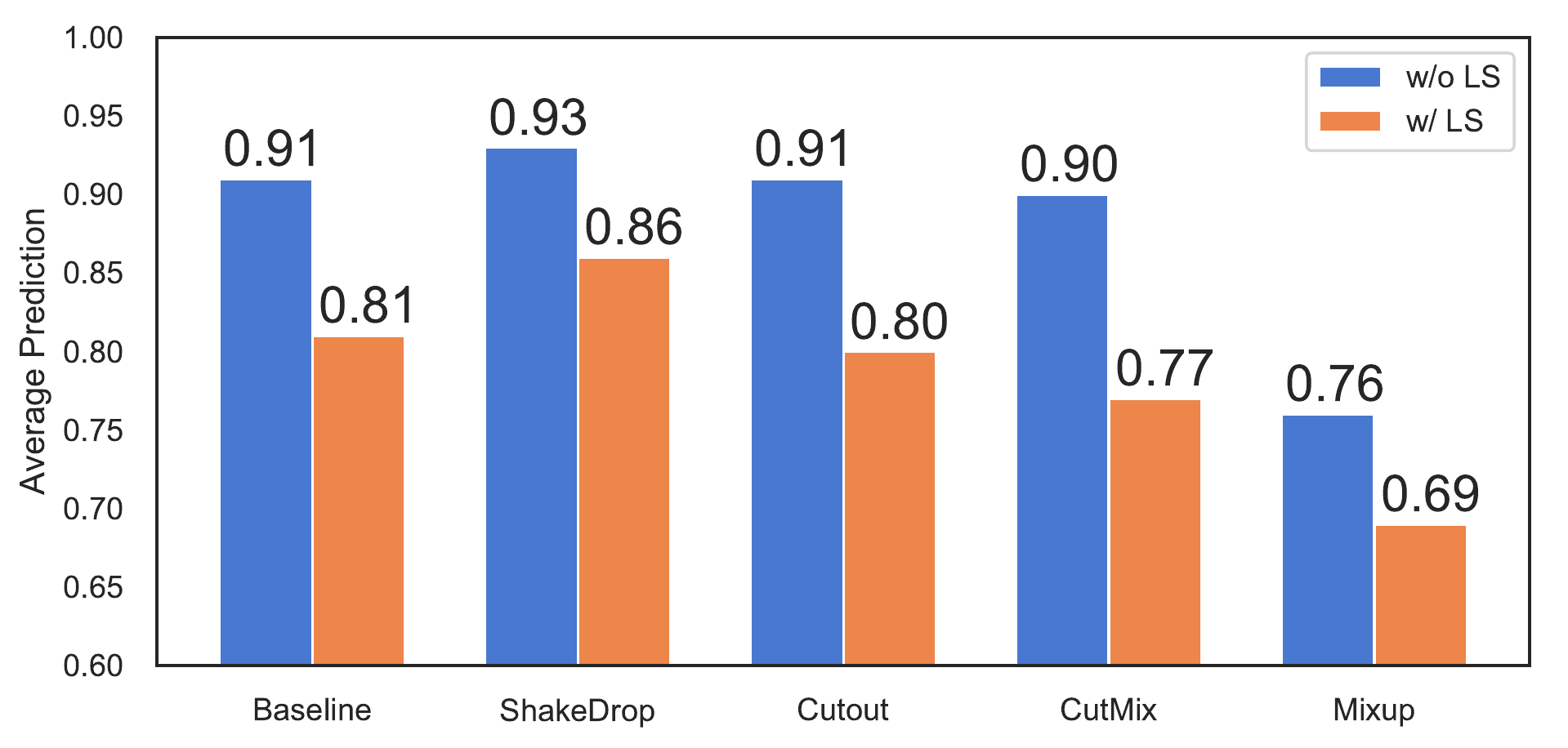}
    \vspace{-0.4cm}
    \caption{Average top-1 prediction probability by models trained with state-of-the-art regularization methods. Models with label smoothing (LS) produce less confident predictions.}
    \label{fig:prediction_ls}
\end{figure}
\vspace{-0.2cm}

\begin{table*}[t!]
\centering
\caption{Top-1 errors of considered regularization tehcniques on various test-time perturbations. We report the average Top-1 error among clean images, FGSM attacked images, occluded images, and naturally corrupted images (ImageNet-C). Finally, we report mCE (mean corruption error) normalized by AlexNet. SD and LS stand for ShakeDrop and label smoothing, respectively.}
\label{table:imagenet}
\begin{tabular}{@{}lc|ccccccc|c@{}}
\toprule
                 & Average & Clean & FGSM & Occ. & Noise  & Blur   & Weather & Digital & mCE    \\ \midrule
Baseline         & 67.43  & 23.68   & 91.85 & 46.01    & 78.58 & 86.63 & 64.99  & 80.24  & 77.55 \\
Label Smoothing  & 62.67  & 22.31   & 73.60 & 44.35    & 77.08 & 82.30 & 61.72  & 77.33  & 74.44 \\
ShakeDrop        & 64.57  & 22.03   & 87.19 & 42.98    & 76.13 & 83.42 & 61.56  & 78.69  & 74.87 \\
ShakeDrop + LS   & 61.45  & 21.92   & 72.65 & 42.85    & 74.47 & 82.15 & 60.47  & 75.67  & 73.10 \\
Cutout           & 64.81  & 22.93   & 88.50 & 29.72    & 79.94 & 85.37 & 65.34  & 81.87  & 78.01 \\
Cutout + LS      & 61.90  & 22.02   & 75.24 & 29.08    & 79.80 & 84.51 & 62.72  & 79.93  & 76.54 \\
Mixup            & 61.46  & 22.58   & 75.60 & 44.20    & 73.09 & 81.49 & 58.83  & 74.42  & 71.88 \\
Mixup + LS       & 58.54  & 22.41   & 69.43 & 42.31    & 65.36 & 82.95 & 53.37  & 73.94  & 69.14 \\
CutMix           & 62.08  & 21.60   & 69.04 & 30.09    & 80.88 & 84.87 & 64.11  & 83.95  & 78.29 \\
CutMix + LS      & 61.02  & 21.87   & 67.41 & 31.51    & 77.01 & 84.61 & 63.13  & 81.56  & 76.55 \\
CutMix + SD      & 61.75  & 21.60   & 80.00 & 31.28    & 77.06 & 84.18 & 61.04  & 77.07  & 74.69 \\
CutMix + SD + LS & 60.96  & 21.90   & 68.65 & 31.62    & 76.04 & 84.53 & 62.82  & 81.16  & 76.14 \\ \bottomrule
\end{tabular}
\end{table*}

\textit{Can label smoothing help adversarial robustness and uncertainty estimates?} In our experiments, adding label smoothing (LS) alone does not generally improve classification accuracies. Surprisingly, however, we observe that LS improves robustness against adversarial perturbation, calibration error, and OOD detection performance (Table~\ref{table:overview}). For example, by adding LS, Cutout + ShakeDrop achieves $13.49\%$ classification top-1 error and FGSM top-1 error $69.59\%$ where performances without LS are $15.91\%$ and $88.66\%$ for classification and adversarial robustness respectively. We believe that it is because a model trained with LS produces low confident predictions in general (Figure~\ref{fig:prediction_ls}). In particular, LS shows impressive improvements in the expected calibration error, except for Mixup and CutMix families. We believe the result is due to the fact that Mixup and CutMix already contain the label mixing stage that already lowers the prediction confidences; further adding label smoothing makes the overall confidences too low.

\textit{Can well-regularized models be a powerful baseline for the robustness and uncertainty estimations?} In Table~\ref{table:powerful_baseline}, we have observed that our well-regularized models such as Cutout + ShakeDrop + label smoothing, Mixup + ShakeDrop + label smoothing, and CutMix + ShakeDrop outperform methods targeted for improved robustness and uncertainty estimations (ALP and OOD augmentations) in many evaluation metrics. For example, ALP model shows occlusion top-1 error $92.27\%$ while Cutout and CutMix based models show $26.33\%$ and $34.96\%$ top-1 error respectively. It is notable that OOD augmentations are not effective for CIFAR-100 tasks, while they have been shown to be effective for toy datasets like SVHN and CIFAR-10~\cite{lee2018confident}.

\subsection{ImageNet Experiments}

In this section, we report experimental results on ImageNet. We use ResNet-50 \cite{resnet} as the baseline model and train the models with same training scheme as used in \cite{yun2019cutmix}. We only evaluate robustness benchmarks, i.e., adversarial robustness against FGSM, natural robustness against ImageNet-C, and robustness to occlusion.

In Table~\ref{table:imagenet}, we report the top-1 error on clean images, attacked images, occluded images, naturally corrupted images (subsets of ImageNet-C), and their average. Also we report the mCE (mean corrupted error) normalized by AlexNet \cite{alexnet} which is proposed in \cite{hendrycks2018imagenet-c}.

As we observed in CIFAR-100 experiments, regularized models provide better overall performances. For example, CutMix achieves $62.08\%$ average error alone but adding ShakeDrop and LS improves average error to $60.96\%$. Table~\ref{table:imagenet} also shows that label smoothing is still effective in improving the robustness of the models in ImageNet experiments. Mixup helps robustness against common corruptions; CutMix shows better classification performance, adversarial robustness, and occlusion robustness.

Interestingly, in our experiments, Mixup + label smoothing achieves the state-of-the-art performance on ImageNet-C mCE of $69.14\%$ where current best model is stylized-ImageNet trained model \cite{geirhos2018imagenet_stylize} with mCE of $69.3\%$. Note that stylized-ImageNet requires heavy pre-computations to generate the stylized images, and requires additional fine-tuning on ImageNet data.

Methods used in our experiments improve the overall robustness and uncertainty performances at negligible additional costs. We believe that well-regularized models should be considered as powerful baselines for the robustness and the uncertainty estimation benchmarks.

\section{Conclusion} 
In this paper, we have empirically compared the robustness and uncertainty estimates of state-of-the-art regularization methods against prior methods specifically designed for such aspects. We have observed that methods proposed to solve the specific problem are only effective on their targeted task. For example, adversarial training only improves adversarial robustness while it degrades classification performance, robustness against common corruptions and occlusion, and uncertainty estimates. On the other hand, good combinations of simple and cheap regularization techniques improve overall robustness and uncertainty estimation performances, and even surpass specialized methods in certain uncertainty and robustness tasks. We believe that well-regularized models have largely been overlooked in robustness and uncertainty studies, and that they should be considered as powerful baselines in future works.

\bibliography{reference}
\bibliographystyle{icml2019}

\end{document}